\documentclass[article, nojss]{jss}\usepackage[]{graphicx}\usepackage[]{xcolor}
\makeatletter
\def\maxwidth{ %
  \ifdim\Gin@nat@width>\linewidth
    \linewidth
  \else
    \Gin@nat@width
  \fi
}
\makeatother

\definecolor{fgcolor}{rgb}{0.345, 0.345, 0.345}

\usepackage{framed}
\makeatletter
 {\par\unskip\endMakeFramed%
 \at@end@of@kframe}
\makeatother

\definecolor{shadecolor}{rgb}{.97, .97, .97}
\definecolor{messagecolor}{rgb}{0, 0, 0}
\definecolor{warningcolor}{rgb}{1, 0, 1}
\definecolor{errorcolor}{rgb}{1, 0, 0}

\usepackage{alltt}

\usepackage{thumbpdf,lmodern}

\usepackage{framed}
\usepackage{amsmath,amsthm}
\usepackage{amsfonts}
\usepackage{amssymb}
\usepackage{rotating}
\usepackage{orcidlink}
\usepackage{enumerate}
\usepackage{float} 
\usepackage{longtable}
\usepackage[tableposition=below]{caption}
\usepackage{subcaption}

\usepackage{lmodern}

\theoremstyle{definition}

\newcommand{\xv}{\mathbf{x}}

\usepackage{listings}
\definecolor{backgroundCol}{rgb}{.97, .97, .97}
\definecolor{commentstyleCol}{rgb}{0.678,0.584,0.686}
\definecolor{keywordstyleCol}{rgb}{0.737,0.353,0.396}
\definecolor{stringstyleCol}{rgb}{0.192,0.494,0.8}
\definecolor{NumCol}{rgb}{0.686,0.059,0.569}
\definecolor{basicstyleCol}{rgb}{0.345, 0.345, 0.345}       
\usepackage{amsmath}
\usepackage{amsfonts}

\let\proglang=\textsf
\let\inlinecode=\texttt

\usepackage{cleveref}

\usepackage{etoolbox}
\usepackage{verbatim,color,lipsum}

\newenvironment{codeoutput}
{\small\color{black}\verbatim}
{\endverbatim}

\definecolor{ccellcolor}{HTML}{d09000}

\newcount\ccellA
\newcommand*\ccell[2]
{%
	\def\tmpa{}%
	\ccellA=1
	\loop
	\ifnum#1=\ccellA
	\edef\tmpa{\unexpanded\expandafter{\tmpa\cellcolor{ccellcolor}}}%
	\fi
	\ifnum#2>\ccellA
	\advance\ccellA1
	\edef\tmpa{\unexpanded\expandafter{\tmpa&}}%
	\repeat
	\tmpa
}

\usepackage{wrapfig}

\author{Susanne Dandl~\orcidlink{0000-0003-4324-4163}\\
LMU Munich  \\
MCML\\
   \And Marc Becker~\orcidlink{0000-0002-8115-0400}\\
   LMU Munich \\
   \And Bernd Bischl~\orcidlink{0000-0001-6002-6980} \\
   LMU Munich \\
   MCML\\
   \AND Giuseppe Casalicchio~\orcidlink{0000-0001-5324-5966} \\
   LMU Munich \\
   MCML
   \And Ludwig Bothmann~\orcidlink{0000-0002-1471-6582} \\
	LMU Munich \\
	MCML}
\Plainauthor{Susanne Dandl, Marc Becker, Bernd Bischl, Giuseppe Casalicchio, Ludwig Bothmann}

\title{\pkg{mlr3summary}: Concise and interpretable summaries for machine learning models}
\Plaintitle{mlr3summary: Concise and interpretable summaries for machine learning models}
\Shorttitle{\pkg{mlr3summary}: Concise and interpretable summaries for machine learning models}

\Abstract{
   This work introduces a novel \proglang{R} package for concise, informative summaries of machine learning models. 
   We take inspiration from the summary function for (generalized) linear models in \proglang{R}, but extend it in several directions: 
   First, our summary function is model-agnostic and provides a unified summary output also for non-parametric machine learning models; 
   Second, the summary output is more extensive and customizable -- it comprises information on the dataset, model performance, model complexity, model's estimated feature importances, feature effects, and fairness metrics; 
   Third, models are evaluated based on resampling strategies for unbiased estimates of model performances, feature importances, etc.
   Overall, the clear, structured output should help to enhance and expedite the model
   selection process, making it a helpful tool for practitioners and researchers alike.
}

\Keywords{Model summary, interpretable machine learning, resampling-based evaluation} 
\Plainkeywords{Model summary, interpretable machine learning, resampling-based evaluation}

\Address{
  Ludwig Bothmann \\
  Institut f\"ur Statistik \\ Ludwig-Maximilians-Universit\"at M\"unchen, Germany \\
  Ludwigstr.\ 33, 80539 Munich, Germany \\
  Munich Center for Machine Learning (MCML), Germany \\
  E-mail: \email{Ludwig.Bothmann@stat.uni-muenchen.de}
}
\IfFileExists{upquote.sty}{\usepackage{upquote}}{}
\begin{document}



\section{Introduction}
\label{sec:intro}

Machine learning (ML) increasingly supports decision-making processes in various domains. 
A data scientist has a wide range of models available, ranging from intrinsically interpretable models such as linear models to highly complex models such as random forests or gradient boosted trees.
Intrinsically interpretable models can come at the expense of generalization performance, i.e., the model's capability to predict accurately on future data. 
Being able to interpret predictive models is either often a strict requirement for scientific inference or at least a very desirable property to audit models in other (more technical) contexts. Many methods have been proposed for interpreting black-box ML models in the field of interpretable ML (IML).

For comparing (generalized) linear models (GLMs), the \pkg{stats} package in \proglang{R} offers a \inlinecode{summary} function, which only requires the model (fitted with \inlinecode{lm} or \inlinecode{glm}) as input.  
As an example, \inlinecode{glm} is applied to a preprocessed version of the German credit dataset \citep{creditdata} (available in the package via \inlinecode{data("credit", package = "mlr3summary")}): 

\begin{lstlisting}[language=R]
> logreg = glm(risk ~., data = credit, 
+   family = binomial(link = "logit"))
> summary(logreg)
\end{lstlisting}
\begin{codeoutput}
	Call:
	glm(formula = risk ~., data = credit, family = binomial(link = "logit"))
	Coefficients:
	Estimate Std. Error z value Pr(>|z|)    
	(Intercept)        1.057e+00  3.646e-01   2.900  0.00373 ** 
	age                9.103e-03  8.239e-03   1.105  0.26925    
	...
	Residual deviance: 656.19  on 515  degrees of freedom
	AIC: 670.19
	...
\end{codeoutput}
% ...
%
This (shortened) summary informs about the significance of variables  (\inlinecode{Pr(>|z|)}), their respective effect size and direction (\inlinecode{Estimate}), as well as the goodness-of-fit of the model (\inlinecode{Residual deviance} and \inlinecode{AIC}).
Unfortunately, many other non-parametric ML models currently cannot be analyzed similarly: either targeted implementations exist for specific model classes, or an array of different model-agnostic interpretability techniques (e.g., to derive feature importance) scattered across multiple packages \citep{iml,dalex,explainer} must be employed. 
However, especially in applied data science, a user often performs model selection or model comparison across an often diverse pool of candidate models, so a standardized diagnostic output becomes highly desirable. 

Another issue is that in the \texttt{glm}-based \texttt{summary}, the goodness-of-fit is only evaluated on the training data, but not on hold-out/test data. 
While this might be appropriate for GLM-type models -- provided proper model diagnosis has been performed -- this is not advisable for non-parametric and non-linear models, which can overfit the training data.\footnote{For completeness' sake: Overfitting \textit{can} happen for GLMs, e.g., in high-dimensional spaces with limited sample size.}
Here, hold-out test data or in general resampling techniques like cross-validation should be used for proper estimation of the generalization performance \cite{simon_resampling_2007}. 
Such resampling-based performance estimation should also be used for loss-based IML methods. 
For interpretability methods that only rely on predictions, this might also be advisable but might not lead to huge differences in results \citep{molnar_general_2022, molnar_relating_2023}.

\paragraph{Contributions}
With the \pkg{mlr3summary} package, we provide a novel model-agnostic \inlinecode{summary} function for ML models and learning algorithms in \proglang{R}.
This is facilitated by building upon \pkg{mlr3} \citep{mlr3,mlr3book} -- a package ecosystem for applied ML, including resampling-based performance assessment.
The \inlinecode{summary} function returns a structured overview that gives information on the underlying dataset and model, generalization performances, complexity of the model, fairness metrics, and feature importances and effects.
For the latter two, the function relies on model-agnostic methods from the field of IML.
The output is customizable via a flexible \inlinecode{control} argument to allow adaptation to different application scenarios.
The \pkg{mlr3summary} package is released under LGPL-3 on GitHub (\url{https://github.com/mlr-org/mlr3summary}) and CRAN (\url{https://cran.r-project.org/package=mlr3summary}). 
Documentations in the form of help pages are available as well as unit tests. 
The example code of this manuscript is available via \inlinecode{demo("credit", package = "mlr3summary")}.

\section{Related work}
\label{sec:rel-work}

Most \proglang{R} packages that offer model summaries are restricted to parametric models and extend the \pkg{stats} \texttt{summary} method (e.g., \pkg{modelsummary} \citep{modelsummary}, \pkg{broom} \citep{broom}).
Performance is only assessed based on training data -- generalization errors are not provided.
Packages that can handle diverse ML models focus primarily on performance assessment (e.g., \pkg{mlr3} \citep{mlr3}, \pkg{caret} \citep{caret}).
Packages that primarily consider feature importances and effects do not provide overviews in a concise, decluttered format but provide extensive reports (e.g., \pkg{modelDown} \citep{modelDown} and \pkg{modelStudio} \citep{modelStudio} based on \pkg{DALEX} \citep{dalex}, or \pkg{explainer} \citep{explainer}). 
While it is possible to base the assessment on hold-out/test data, assessment based on resampling is not automatically supported by these packages. 
Overall, to the best of our knowledge, there is no  \proglang{R} package yet that allows for a concise yet informative overview based on resampling-based performance assessment, model complexity, feature importance and effect directions, and fairness metrics.

\section{Design, functionality, and example}

The core function of the \pkg{mlr3summary} package is the \inlinecode{S3}-based \inlinecode{summary} function for \pkg{mlr3} \inlinecode{Learner} objects. 
It has three arguments: \inlinecode{object} reflects a trained model -- a model of class \inlinecode{Learner} fitted with \pkg{mlr3};  \inlinecode{resample\_result} reflects the results of resampling -- a \inlinecode{ResampleResult} object fitted with \pkg{mlr3}; \inlinecode{control} reflects some control arguments -- a list created with \inlinecode{summary\_control} (details in \Cref{subsec:customize}).

The \pkg{mlr3} package is the basis of \pkg{mlr3summary}  because it provides a unified interface to diverse ML models and resampling strategies.
A general overview of the \pkg{mlr3} ecosystem is given in \citeauthor{mlr3book} \citep{mlr3book}. 
With \pkg{mlr3}, the modelling process involves the following steps: (1) initialize a regression or classification task, (2) choose a regression or classification learner, (3) train a model with the specified learner on the initialized task, (4) apply a resampling strategy.
The last step is necessary to receive valid estimates for  performances, importances, etc., as mentioned in \Cref{sec:intro}.
The following lines of code illustrate steps (1)-(4) on the (preprocessed) credit dataset from \Cref{sec:intro} using a ranger random forest.
As a resampling strategy, we conduct $3$-fold cross-validation. 

\begin{lstlisting}[language=R]
> task = TaskClassif$new(id = "credit", backend = credit, 
+   target = "risk")
> rf = lrn("classif.ranger", predict_type = "prob")
> rf$train(task)
> cv3 = rsmp("cv", folds = 3L)
> rr = resample(task = task, learner = rf, resampling = cv3, 
+   store_models = TRUE)
\end{lstlisting}

Internally, the \inlinecode{resample} function fits, in each iteration, the model on the respective training data, uses the model to predict the held-out test data, and stores the predictions in the result object.
To compute performances, complexities, importances, and other metrics, the \inlinecode{summary} function iteratively accesses the models and datasets within the resulting resample object, which requires setting the parameter \inlinecode{store\_models = TRUE} within the \inlinecode{resample} function.
For the final \inlinecode{summary} output, the results of each iteration are aggregated (e.g., averages and standard deviations (sds)).

\subsection{Summary function and output}
This section shows the \inlinecode{summary} call and output for the random forest of the previous credit example and provides some details on each displayed paragraph. 

\begin{lstlisting}[language=R]
> summary(object = rf, resample_result = rr)
\end{lstlisting}
\includegraphics[width = 1\textwidth, trim = {0 0 5cm 0}, clip]{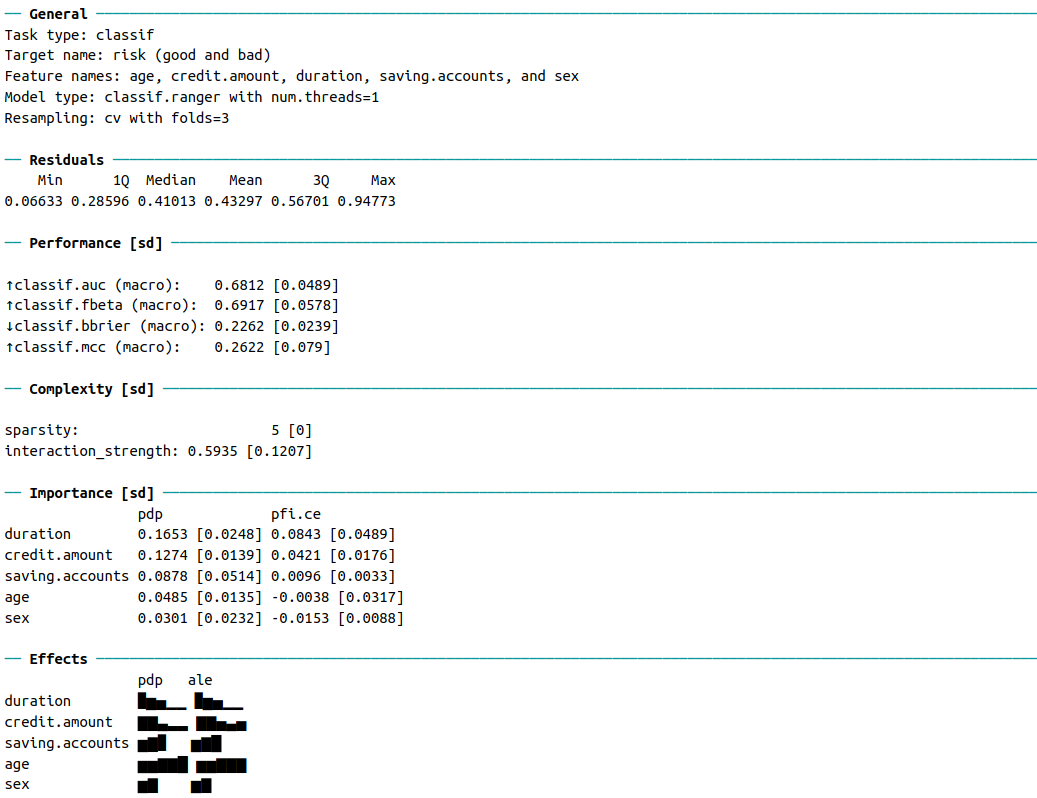}
\textbf{General} provides an overview of the task, the learner (including its hyperparameters), and the resampling strategy.\footnote{Currently, this is the only paragraph that is based on \inlinecode{object}, all other paragraphs are based on \inlinecode{resample\_result}.} 

\textbf{Residuals} display the distribution of residuals of \textit{hold-out} data over the resampling iterations. 
For regression models, the residuals display the difference between true and predicted outcome.
For classifiers that return class probabilities, the residuals are defined as the difference between predicted probabilities and a one-hot-encoding of the true class.
For classifiers that return classes, a confusion matrix is shown.

\textbf{Performance} displays averages and sds (in $[\,]$) of performance measures over the iterations.\footnote{Please note that there is no unbiased estimator of the variance, see  \citep{nadeau_ge_1999} and \Cref{sec:outlook} for a discussion.}
The shown performance values are the area-under-the-curve (auc), the F-score (fbeta), the binary Brier score (bbrier), and Mathew's correlation coefficient (mcc). 

The arrows display whether lower or higher values refer to a better performance.
``(macro)'' indicates a macro aggregation, i.e., measures are computed for each iteration separately before averaging. ``(micro)'' would indicate that measures are computed across all iterations (see \cite{mlr3book} for details).

\textbf{Complexity} displays averages and sds of two model complexity measures proposed by \citeauthor{molnar_complexity_2020} \citep{molnar_complexity_2020}:
\inlinecode{sparsity} shows the number of used features that have a non-zero effect on the prediction (evaluated by accumulated local effects (ale) \citep{apley_ale_2020}); \inlinecode{interaction\_strength} shows the scaled approximation error between a main effect model (based on ale) and the prediction function.\footnote{The interaction strength has a value in $[0, 1]$, 0 means no interactions, 1 means no main effects but interactions.}

\textbf{Importance} shows the averages and sds of feature importances over the iterations.
The first column (pdp) displays importances based on the sds of partial dependence curves \citep{friedman_pdp_2001,greenwell_pdpfi_2018}, the second column (pfi.ce) shows the results for permutation feature importance \cite{breiman_pfi_2001, rudin_pfi_2018}.

\textbf{Effects} shows average effect plots over the iterations --  partial dependence plots (pdp) and ale plots \citep{friedman_pdp_2001,apley_ale_2020}.
For binary classifiers, the effect plots are only shown for the positively-labeled class (here, \inlinecode{task\$positive = "good"}).
For multi-class classifiers, the effect plots are given for each outcome class separately (one vs.\ all).
For categorical features, the bars are ordered according to the factor levels of the feature.

The learner can also be a complete pipeline from \pkg{mlr3pipelines} \citep{mlr3pipelines}, where the most common case would be an ML model with associated pre-processing steps.
Then, the \inlinecode{summary} output also shows some basic information about the pipeline.\footnote{Linear pipelines can be displayed in the console, non-linear parts are suppressed in the output.} 
Since preprocessing steps are treated as being part of the learner, the summary output is displayed on the original data (e.g., despite one-hot encoding of categorical features, importance results are not shown for each encoding level separately).
The learner can also be an AutoTuner from \pkg{mlr3tuning}, where automatic processes for tuning the hyperparameters are conducted.
Examples on pipelining and tuning are given in the demo of the package.

\subsection{Customizations}
\label{subsec:customize}

The output of the \inlinecode{summary} function can be customized via a \inlinecode{control} argument which requires a list created with the function \inlinecode{summary\_control} as an input. 
If no control is specified, the following default setting is used:

\begin{lstlisting}[language=R]
> summary_control(measures = NULL, 
+   complexity_measures = c("sparsity", "interaction_strength"),
+   importance_measures = NULL, n_important = 15L,
+   effect_measures = c("pdp", "ale"),
+   fairness_measures = NULL, protected_attribute = NULL,
+   hide = NULL, digits = max(3L, getOption("digits") - 3L))
\end{lstlisting}
Performances are adaptable via \inlinecode{measures}, complexities via \inlinecode{complexity\_measures}, importances via \inlinecode{importance\_measures} and effects via \inlinecode{effect\_measures} within \inlinecode{summary\_control}. 
Examples are given in the demo of the package.
The default for \inlinecode{measures} and \inlinecode{importance\_ measures} is \inlinecode{NULL}, which results in a collection of commonly reported measures being chosen, based on the task type -- for concrete measures see the help page (\inlinecode{?summary\_control}).
\inlinecode{n\_important} reflects that, by default, only the $15$ most important features are displayed in the output.
This is especially handy for high-dimensional data.
With \inlinecode{hide}, paragraphs of the summary output can be omitted (e.g., \inlinecode{"performance"}) and with \inlinecode{digits}, the number of printed digits is specified. 

Fairness assessment for classification and regression models is also available in \pkg{mlr3summary} based on the \pkg{mlr3fairness} package \citep{mlr3fairness}. 
Therefore, a protected attribute must be specified. 
This can be done either within the task by updating the feature roles or by specifying a \inlinecode{protected\_attribute} in \inlinecode{summary\_control}.
The following shows the code and output when specifying \inlinecode{sex} as a protected attribute. The shown default fairness measures are demographic parity (dp), conditional use accuracy equality (cuae) and equalized odds (eod), other measures are possible via \inlinecode{fairness\_measures} in \inlinecode{summary\_control}.

\begin{lstlisting}[language=R]
> summary(object = rf, resample_result = rr, 
+   control = summary_control(protected_attribute = "sex"))
\end{lstlisting}
\includegraphics[width = 1\textwidth, trim = {0 0 0cm 0}, clip]{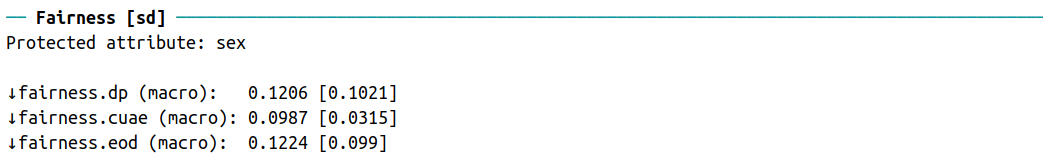}

\section{Runtime assessment}
\begin{figure}[ht]
	\centering
	\includegraphics[width = 1\textwidth]{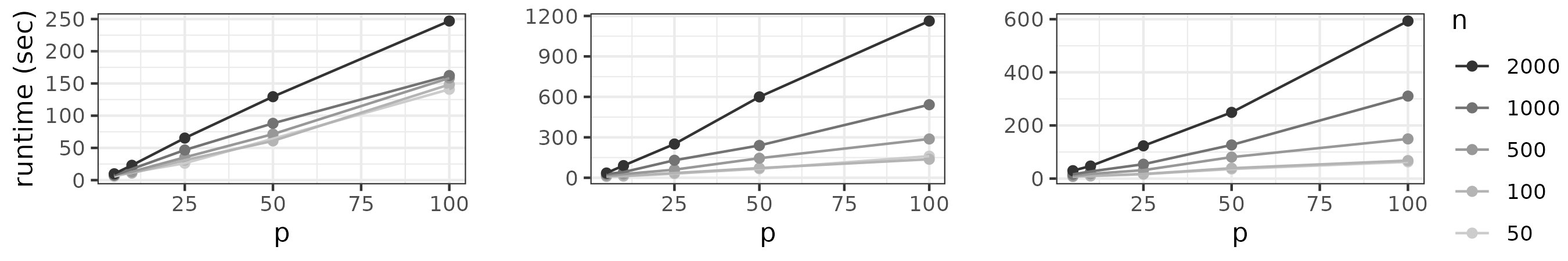}
	\caption{Runtimes of the \inlinecode{summary} function for linear models (left), and random forests without (middle) and with (right) parallelization, for differing numbers of features $p$ and observations $n$.}
	\label{fig:runtimes}
\end{figure}

To assess how the runtime scales with differing numbers of features $p \in \{5, 10, 25, 50, 100\}$ and numbers of observations $n \in \{50, 100, 500, 1000, 2000\}$, we conducted a simulation study. Given $X_1, X_2, X_3 \sim U(0, 1)$, $X_4 \sim Bern(0.75)$, the data generating process 
is $y = f(\xv) + \epsilon$ with $f(\xv) = 4x_1 + 4x_2 + 4x_4x_3^2$ and $\epsilon \sim N(0, 0.1\cdot f(\xv))$. 
As noise variables, $X_5$ as a categorical feature with five classes, and $X_6, ..., X_{p} \sim N(0, 1)$ were added to the data. 
We trained random forests and linear main effect models on the datasets and conducted 3-fold cross-validation. The first two figures in \Cref{fig:runtimes} show that runtimes of the linear model were lower compared to the random forest. 
To improve runtimes, we added parallelization over the resampling iterations (via the \pkg{future} package \citep{future}) as another feature to \pkg{mlr3summary} -- results for the random forest (with $3$ cores) are on the right.
Overall, scaling of runtimes is worse in $p$ than in $n$.

\section{Outlook and discussion}
\label{sec:outlook}

In conclusion, this paper introduces a novel \proglang{R} package for concise model summaries. 
The summary output is highly adaptable due to a \inlinecode{control} argument and might be extended in the future.
We also plan to offer a \inlinecode{report} function for detailed visualizations and model comparisons.
To assess importance and effects of single features, \pkg{mlr3summary} builds upon the \pkg{iml} and \pkg{fastshap} packages.
These packages only offer a limited set of interpretation methods. 
Recommended alternatives to permutation feature importances like conditional feature importance \cite{molnar_general_2022}, are currently not available in a proper \proglang{R} package (published on CRAN).
Our summary also currently lacks proper statistical tests for importances or confidence intervals for performances. This is because unbiased estimates of the variance are required which is a challenge for resampling strategies and the available methods that propose unbiased estimates are computationally infeasible (e.g., due to many model refits) \citep{molnar_relating_2023,bates_cv_2023}.
Addressing this issue requires some concerted efforts from the research community. 
If methods are readily available in \proglang{R}, we are happy to integrate them in \pkg{mlr3summary}. 

\section*{Acknowledgments}
	This work has been partially supported by the Federal Statistical Office of Germany.

\bibliography{literature}

\begin{thebibliography}{25}
\newcommand{\enquote}[1]{``#1''}
\providecommand{\natexlab}[1]{#1}
\providecommand{\url}[1]{\texttt{#1}}
\providecommand{\urlprefix}{URL }
\expandafter\ifx\csname urlstyle\endcsname\relax
  \providecommand{\doi}[1]{doi:\discretionary{}{}{}#1}\else
  \providecommand{\doi}{doi:\discretionary{}{}{}\begingroup
  \urlstyle{rm}\Url}\fi
\providecommand{\eprint}[2][]{\url{#2}}

\bibitem[{Apley and Zhu(2020)}]{apley_ale_2020}
Apley DW, Zhu J (2020).
\newblock \enquote{Visualizing the Effects of Predictor Variables in Black Box
  Supervised Learning Models.}
\newblock \emph{Journal of the Royal Statistical Society Series B: Statistical
  Methodology}, \textbf{82}(4), 1059--1086.
\newblock \doi{10.1111/rssb.12377}.

\bibitem[{Arel-Bundock(2022)}]{modelsummary}
Arel-Bundock V (2022).
\newblock \enquote{{modelsummary}: Data and Model Summaries in {R}.}
\newblock \emph{Journal of Statistical Software}, \textbf{103}(1), 1–23.
\newblock \doi{10.18637/jss.v103.i01}.

\bibitem[{Baniecki and Biecek(2019)}]{modelStudio}
Baniecki H, Biecek P (2019).
\newblock \enquote{{modelStudio}: {I}nteractive Studio with Explanations for
  {ML} Predictive Models.}
\newblock \emph{Journal of Open Source Software}, \textbf{4}(43), 1798.
\newblock \doi{10.21105/joss.01798}.

\bibitem[{Bengtsson(2021)}]{future}
Bengtsson H (2021).
\newblock \enquote{A Unifying Framework for Parallel and Distributed Processing
  in R using Futures.}
\newblock \emph{The R Journal}, \textbf{13}(2), 208--227.
\newblock \doi{10.32614/RJ-2021-048}.

\bibitem[{Biecek(2018)}]{dalex}
Biecek P (2018).
\newblock \enquote{{DALEX}: Explainers for Complex Predictive Models in {R}.}
\newblock \emph{Journal of Machine Learning Research}, \textbf{19}(84), 1--5.

\bibitem[{Binder \emph{et~al.}(2021)Binder, Pfisterer, Lang, Schneider,
  Kotthoff, and Bischl}]{mlr3pipelines}
Binder M, Pfisterer F, Lang M, Schneider L, Kotthoff L, Bischl B (2021).
\newblock \enquote{{mlr3pipelines} - {F}lexible Machine Learning Pipelines in
  {R}.}
\newblock \emph{Journal of Machine Learning Research}, \textbf{22}(184), 1--7.

\bibitem[{Bischl \emph{et~al.}(2024)Bischl, Sonabend, Kotthoff, and
  Lang}]{mlr3book}
Bischl B, Sonabend R, Kotthoff L, Lang M (2024).
\newblock \emph{Applied machine learning using {mlr3} in {R}}.
\newblock Chapman and Hall/CRC.
\newblock ISBN 9781003402848.
\newblock \doi{10.1201/9781003402848}.

\bibitem[{Breiman(2001)}]{breiman_pfi_2001}
Breiman L (2001).
\newblock \enquote{Random Forests.}
\newblock \emph{Machine Learning}, \textbf{45}(1), 5–32.
\newblock ISSN 0885-6125.
\newblock \doi{10.1023/a:1010933404324}.

\bibitem[{Fisher \emph{et~al.}(2019)Fisher, Rudin, and
  Dominici}]{rudin_pfi_2018}
Fisher A, Rudin C, Dominici F (2019).
\newblock \enquote{All Models are Wrong, but Many are Useful: Learning a
  Variable's Importance by Studying an Entire Class of Prediction Models
  Simultaneously.}
\newblock \emph{Journal of Machine Learning Research}, \textbf{20}(177).

\bibitem[{Friedman(2001)}]{friedman_pdp_2001}
Friedman JH (2001).
\newblock \enquote{Greedy Function Approximation: A Gradient Boosting Machine.}
\newblock \emph{The Annals of Statistics}, \textbf{29}(5).
\newblock ISSN 0090-5364.
\newblock \doi{10.1214/aos/1013203451}.

\bibitem[{Greenwell \emph{et~al.}(2018)Greenwell, Boehmke, and
  McCarthy}]{greenwell_pdpfi_2018}
Greenwell BM, Boehmke BC, McCarthy AJ (2018).
\newblock \enquote{A Simple and Effective Model-Based Variable Importance
  Measure.}
\newblock \emph{arXiv preprint arXiv:1805.04755}.
\newblock \doi{10.48550/arXiv.1805.04755}.

\bibitem[{Hofmann(1994)}]{creditdata}
Hofmann H (1994).
\newblock \enquote{Statlog ({G}erman Credit Data).}
\newblock \emph{{UCI} Machine Learning Repository}.
\newblock \doi{10.24432/C5NC77}.

\bibitem[{{Kuhn} and {Max}(2008)}]{caret}
{Kuhn}, {Max} (2008).
\newblock \enquote{Building Predictive Models in {R} Using the {caret}
  Package.}
\newblock \emph{Journal of Statistical Software}, \textbf{28}(5), 1–26.
\newblock \doi{10.18637/jss.v028.i05}.

\bibitem[{Lang \emph{et~al.}(2019)Lang, Binder, Richter, Schratz, Pfisterer,
  Coors, Au, Casalicchio, Kotthoff, and Bischl}]{mlr3}
Lang M, Binder M, Richter J, Schratz P, Pfisterer F, Coors S, Au Q, Casalicchio
  G, Kotthoff L, Bischl B (2019).
\newblock \enquote{{mlr3}: A Modern Object-oriented Machine Learning Framework
  in {R}.}
\newblock \emph{Journal of Open Source Software}.
\newblock \doi{10.21105/joss.01903}.

\bibitem[{Molnar \emph{et~al.}(2018)Molnar, Bischl, and Casalicchio}]{iml}
Molnar C, Bischl B, Casalicchio G (2018).
\newblock \enquote{{iml}: An {R} Package for Interpretable Machine Learning.}
\newblock \emph{JOSS}, \textbf{3}(26), 786.
\newblock \doi{10.21105/joss.00786}.

\bibitem[{Molnar \emph{et~al.}(2020)Molnar, Casalicchio, and
  Bischl}]{molnar_complexity_2020}
Molnar C, Casalicchio G, Bischl B (2020).
\newblock \emph{Quantifying model complexity via functional decomposition for
  better post-hoc interpretability}, p. 193–204.
\newblock Springer International Publishing.
\newblock \doi{10.1007/978-3-030-43823-4_17}.

\bibitem[{Molnar \emph{et~al.}(2023)Molnar, Freiesleben, K{\"o}nig, Herbinger,
  Reisinger, Casalicchio, Wright, and Bischl}]{molnar_relating_2023}
Molnar C, Freiesleben T, K{\"o}nig G, Herbinger J, Reisinger T, Casalicchio G,
  Wright MN, Bischl B (2023).
\newblock \enquote{Relating the Partial Dependence Plot and Permutation Feature
  Importance to the Data Generating Process.}
\newblock In L~Longo (ed.), \emph{Explainable Artificial Intelligence}, pp.
  456--479. Springer Nature Switzerland, Cham.
\newblock \doi{10.1007/978-3-031-44064-9_24}.

\bibitem[{Molnar \emph{et~al.}(2022)Molnar, König, Herbinger, Freiesleben,
  Dandl, Scholbeck, Casalicchio, Grosse-Wentrup, and
  Bischl}]{molnar_general_2022}
Molnar C, König G, Herbinger J, Freiesleben T, Dandl S, Scholbeck CA,
  Casalicchio G, Grosse-Wentrup M, Bischl B (2022).
\newblock \enquote{General Pitfalls of Model-Agnostic Interpretation Methods
  for Machine Learning Models.}
\newblock In A~Holzinger, et~al (eds.), \emph{{xxAI} - Beyond Explainable AI:
  International Workshop}, pp. 39--68. Springer International Publishing, Cham.
\newblock ISBN 978-3-031-04083-2.
\newblock \doi{10.1007/978-3-031-04083-2_4}.

\bibitem[{Nadeau and Bengio(1999)}]{nadeau_ge_1999}
Nadeau C, Bengio Y (1999).
\newblock \enquote{Inference for the Generalization Error.}
\newblock In S~Solla, et~al (eds.), \emph{Advances in Neural Information
  Processing Systems}, volume~12, pp. 1--7. MIT Press.

\bibitem[{Pfisterer \emph{et~al.}(2023)Pfisterer, Siyi, and
  Lang}]{mlr3fairness}
Pfisterer F, Siyi W, Lang M (2023).
\newblock \emph{mlr3fairness: Fairness auditing and debiasing for 'mlr3'}.
\newblock {R} package version 0.3.2,
  \urlprefix\url{https://CRAN.R-project.org/package=mlr3fairness}.

\bibitem[{Robinson \emph{et~al.}(2023)Robinson, Hayes, and Couch}]{broom}
Robinson D, Hayes A, Couch S (2023).
\newblock \emph{{broom}: Convert statistical objects into tidy tibbles}.
\newblock {R} package version 1.0.5,
  \urlprefix\url{https://CRAN.R-project.org/package=broom}.

\bibitem[{Romaszko \emph{et~al.}(2019)Romaszko, Tatarynowicz, Urbański, and
  Biecek}]{modelDown}
Romaszko K, Tatarynowicz M, Urbański M, Biecek P (2019).
\newblock \enquote{{modelDown}: Automated Website Generator with Interpretable
  Documentation for Predictive Machine Learning Models.}
\newblock \emph{Journal of Open Source Software}, \textbf{4}(38).
\newblock \doi{10.21105/joss.01444}.

\bibitem[{Simon(2007)}]{simon_resampling_2007}
Simon R (2007).
\newblock \emph{Resampling strategies for model assessment and selection}, pp.
  173--186.
\newblock Springer US, Boston, MA.
\newblock \doi{10.1007/978-0-387-47509-7_8}.

\bibitem[{Stephen~Bates and Tibshirani(2023)}]{bates_cv_2023}
Stephen~Bates TH, Tibshirani R (2023).
\newblock \enquote{Cross-Validation: What Does It Estimate and How Well Does It
  Do It?}
\newblock \emph{Journal of the American Statistical Association}, pp. 1--12.
\newblock \doi{10.1080/01621459.2023.2197686}.

\bibitem[{{Zargari Marandi}(2023)}]{explainer}
{Zargari Marandi} R (2023).
\newblock \emph{{explainer}: Machine learning model explainer}.
\newblock {R} package version 1.0.0,
  \urlprefix\url{https://CRAN.R-project.org/package=explainer}.

\end{thebibliography}
\end{document}